\documentclass[11pt,a4paper]{article}
\usepackage[blocks]{authblk}
\usepackage[hyperref]{naaclhlt2018}
\usepackage{graphics}
\usepackage{graphicx}
\usepackage{times}
\usepackage{latexsym}
\usepackage{booktabs}
\usepackage{url}
\usepackage{siunitx}
\usepackage{colortbl}
\usepackage[utf8]{inputenc}
\usepackage{setspace}

\aclfinalcopy % Uncomment this line for the final submission
\title{Are Automatic Methods for Cognate Detection Good Enough for Phylogenetic Reconstruction in
Historical Linguistics?}

\author{Taraka Rama$^{\spadesuit}$ ~ Johann-Mattis List$^{\diamondsuit}$ ~ Johannes Wahle$^{\heartsuit}$ ~ Gerhard Jäger$^{\heartsuit}$\\
$^{\spadesuit}$Department of Informatics, University of Oslo, Norway \\
$^{\diamondsuit}$Department of Linguistic and Cultural Evolution, MPI-SHH, Jena, Germany \\
$^{\heartsuit}$Department of Linguistics, University of Tübingen, Germany\\
{\small \tt tarakark@ifi.uio.no, list@shh.mpg.de, \{johannes.wahle,gerhard.jaeger\}@uni-tuebingen.de}
}
\date{}

\begin{document}
\maketitle
\begin{abstract}
%Most pertinent work relies on expert classifications of core-vocabulary lists into cognate
%classes. It is an area of active research how to automatize this. The performance of such
%algorithms is generally evaluated according to machine-learning benchmarks. It is open how
%well the results are suitable for phylogenetic inference.

We evaluate the performance of state-of-the-art algorithms for automatic cognate detection by comparing how useful automatically inferred cognates are for the task of phylogenetic inference compared to classical manually annotated cognate sets. 
Our findings suggest that phylogenies inferred from automated cognate sets come close to phylogenies inferred from expert-annotated ones, although on average, the latter are still superior. 
We conclude that future work on phylogenetic reconstruction can profit much from automatic cognate detection.
Especially where scholars are merely interested in exploring the bigger picture of a language family's phylogeny, algorithms for automatic cognate detection are a useful complement for current research on language phylogenies.
\end{abstract}

\section{Introduction}
The task of \emph{cognate detection}, i.e., the search for genetically related words in different languages, has traditionally been regarded as a task that is barely automatable. 
During the last decades, however, automatic cognate detection approaches since \citet{Covington:96} have been constantly
improved following the work of \citet{kondrak2002algorithms}, both regarding the quality of the inferences \citep{List2017c,Jaeger2017}, and the sophistication of the methods \citep{hauer-kondrak:2011:IJCNLP-2011,rama2016siamese,Jaeger2017}, which have been expanded to account for the detection of partial cognates
\cite{list-lopez-bapteste:2016:P16-2}, language specific sound-transition weights \citep{list:2012:LINGVIS2012} or the search of cognates in whole dictionaries \citep{st2017identifying}. 

Despite the progress, none of the automated cognate detection methods have been used for the purpose of inferring phylogenetic trees using modern Bayesian phylogenetic methods \citep{yang1997bayesian} from computational biology.
Phylogenetic trees are hypotheses of how sets of related languages evolved in time.
They can in turn be used for testing additional hypotheses of language evolution, such as the age of language families \citep{gray2003language,chang2015ancestry}, their spread \citep{bouckaert2012mapping,gray2009language}, the rates of lexical change \citep{greenhill2017evolutionary}, or as a proxy for tasks like cognate detection and linguistic reconstruction \citep{bouchardcote2013}. 
By plotting shared traits on a tree and testing how they could have evolved, trees can even be used to test hypotheses independent from language evolution, such as the universality of typological statements \citep{dunn2011evolved}, or the ancestry of cultural traits \citep{jordan2009matrilocal}.

In the majority of these approaches, scholars infer phylogenetic trees with help of \emph{expert-annotated cognate sets} which serve as input to the phylogenetic software which usually follows a Bayesian likelihood framework. 
Unfortunately, expert cognate judgments are only available for a small number of language families which look back on a long tradition of classical comparative linguistic research \citep{campbell2008language}.
Despite the claims that automatic cognate detection is useful for linguists working on less well studied language families, none of the papers actually tested, if automated cognates can be used instead as well for the important downstream task of Bayesian phylogenetic inference. 
So far, scholars have only tested distance-based approaches to phylogenetic reconstruction \citep{wichmann2010evaluating,rama2013bchap,jager2013phylogenetic}, which employ aggregated linguistic distances computed from string similarity algorithms to infer phylogenetic trees.

In order to test whether automatic cognate detection is useful for phylogenetic inference, we collected multilingual wordlists for five different language families (230 languages, cf. section \ref{subsec:data}) and then applied different cognate detection methods (cf. section \ref{sec:autocog}) to infer cognate sets. We then applied the Bayesian phylogenetic inference procedure (cf. section \ref{sec:bayinf}) to the automated and the expert-annotated cognate sets in order to infer phylogenetic trees. These trees were then evaluated against the \emph{family gold standard trees}, based on external linguistic knowledge \citep{Hammarstroem2017}, using the \emph{Generalized Quartet Distance} (cf. section \ref{subsec:gqd}). The results are provided in table \ref{tab:gqd} and the paper is concluded in section \ref{sec:concl}.

To the best of our knowledge, this is the first study in which the performance of several automatic cognate detection methods on the downstream task of phylogenetic inference is compared. 
While we find that on average the trees inferred from the expert-annotated cognate sets come closer to the gold standard trees, the trees inferred from automated cognate sets come surprisingly close to the trees inferred from the expert-annotated ones.

%Applications of phylogenetic inference for testing hypothesis about time depth of languages, spread of languages, evolution of grammatical features.
%We will test if the cognate clusters inferred by the automated methods can be be supplied as an input to Bayesian phylogenetic inference software to infer high quality trees.
%We evaluate the quality of inferred trees against expert family trees through Generalized Quartet distance. We evaluate the methods on five different language families consisting of about 230 languages.
%The structure of the paper is organized as follows. We discuss the different cognate detection methods, cognate clustering, Bayesian inference, and tree comparison measurees in section \ref{sec:exps}.The results of our  are given in section \ref{sec:results}. Finally, we conclude the paper in section \ref{sec:concl}.

\begin{table}[htb]
\centering
  \begin{tabular}{lp{1.5cm}p{1cm}p{1cm}}
  \toprule
  \textbf{Dataset} & \textbf{Mngs.} & \textbf{Lngs.} & \textbf{AMC} \\
  \midrule
  Austronesian & 210 & 45 & 0.79 \\
  Austro-Asiatic & 200 & 58 & 0.90 \\
  Indo-European & 208 & 42 & 0.95 \\
  Pama-Nyungan & 183 & 67 & 0.89 \\
  Sino-Tibetan & 110 & 64 & 0.91 \\
  \bottomrule
\end{tabular}
  \caption{Datasets used in our study. The second, third, and fourth columns show the number of number of meanings, languages and average mutual coverage for each language family respectively.}
  \label{tab:data}
\end{table}

\section{Materials and Methods}\label{sec:exps}
\subsection{Datasets}\label{subsec:data}
Our wordlists were extracted from publicly available datasets from five different language families:
Austronesian \citep{Greenhill2008}, Austro-Asiatic \citep{Sidwell2015}, Indo-European
\citep{Dunn2012}, Pama-Nyungan \citep{Bowern2012}, and Sino-Tibetan \citep{Peiros2004}. 
In order to
make sure that the datasets were amenable for automatic cognate detection, we had to make sure that
the transcriptions employed are readily recognized, and that the data is sufficient for those
methods which rely on the identification of regular sound correspondences. 
The problem of transcriptions was solved by applying intensive semi-automatic cleaning. 
In order to guarantee an optimal data size, we selected a subset of
languages from each dataset, which would guarantee a high \emph{average mutual coverage} (AMC). AMC
is calculated as the average proportion of words shared by all language pairs in a given dataset. 
All analyses were carried out with version 2.6.2 of LingPy \citep{List2017i}.
Table \ref{tab:data} gives an overview on the number of languages, concepts, and the AMC score for
all datasets.\footnote{In order to allow for an easy re-use of our datasets, we linked all language varieties
to Glottolog \citep{Hammarstroem2017} and all concepts to Concepticon \citep{List2016a}. In addition
to the tabular data formats required to run the analyses with our software tools, we also provide
the data in form of the format specifications suggested by the Cross-Linguistic Data Formats
initiative \citep{Forkel2017a}. Data and source code are provided along with the supplementary
material accompanying this paper.}

\subsection{Automatic Cognate Detection}\label{sec:autocog}
The basic workflow for automatic cognate detection methods applied to multilingual wordlists has been extensively described in the literature \citep{hauer-kondrak:2011:IJCNLP-2011,List2014d}. The workflow can be divided into two major steps: (a)\ word similarity calculation, and (b)\ cognate set partitioning. In the first step, similarity or distance scores for all word pairs in the same concept slot in the data are computed. In the second step, these scores are used to partition the words into sets of presumably related words. Since the second step is a mere clustering task for which many solutions exist, the most crucial differences among algorithms can be noted for step\ (a).
 
For our analysis, we tested six different methods for cognate detection:
The Consonant-Class-Matching (CCM) Method \citep{turchin2010analyzing}, the Normalized Edit
Distance (NED) approach \citep{levenshtein1965binary}, the Sound-Class-Based Aligmnent (SCA) method \citep{List2014d}, the LexStat-Infomap method \citep{List2017c}, the SVM method \citep{Jaeger2017}, and the Online PMI
approach \citep{rama2017fast}.
 
The \textbf{CCM} approach first reduces the size of the alphabets in the phonetic transcriptions by mapping
consonants to \emph{consonant classes} and discarding vowels. Assuming that different sounds which
share the same sound class are likely to go back to the same ancestral sound, words which share the
first two consonant classes are judged to be cognate, while words which differ regarding their first two
classes are regarded as non-cognate. 
 
The \textbf{NED} approach first computes the \emph{normalized edit distance} \citep{Nerbonne:97} for all word pairs in given semantic slot and then clusters the words into cognate sets using a flat version of the UPGMA
algorithm \citep{Sokal1958} and a user-defined threshold of maximal distance among the words. We
follow \citet{List2017c} in setting this threshold to 0.75.

The \textbf{SCA} approach is very similar to NED,
but the pairwise distances are computed with help of the Sound-Class-Based Phonetic Alignment algorithm
\citep{List2014d} which employs an extended sound-class model and a linguistically informed scoring
function. Following \citet{List2017c}, we set the threshold for this approach to 0.45.
 
The \textbf{LexStat-Infomap} method builds on the SCA method by employing the same sound-class model, but individual scoring functions are inferred from the data for each
language pair by applying a permutation method and computing the \emph{log-odds scores}
\citep{Eddy2004} from the expected and the attested distribution of sound matches \citep{List2014d}.
While SCA and NED employ flat UGPMA clustering for step 2 of the workflow, LexStat-Infomap further
uses the Infomap community detection algorithm \citep{rosvall2008maps} to partition the words into
cognate set. Following \citet{List2017c}, we set the threshold for LexStat-Infomap to 0.55.

The \textbf{OnlinePMI} approach \citep{rama2017fast} estimates the sound-pair PMI matrix using the online procedure described in \citet{liang2009online}. The approach starts with an empty PMI matrix and a list of synonymous word pairs from all the language pairs. The approach proceeds by calculating the PMI matrix from alignments calculated for each minibatch of word pairs using the current PMI matrix. Then the calculated PMI matrix for the latest minibatch is combined with the current PMI matrix. This procedure is repeated for a fixed number of iterations. We employ the final PMI matrix to calculate pairwise word similarity matrix for each meaning. In an additional step, the similarity score was transformed into a distance score using the sigmoid transformation: $1.0-(1+\exp(-x))^{-1}$ The word distance matrix is then supplied as an input to the Label Propagation algorithm \citep{raghavan2007near} to infer cognate clusters. We set the threshold for the algorithm to be 0.5. 

For the \textbf{SVM} approach \citep{Jaeger2017} a linear SVM classifier was trained with PMI
similarity \citep{jager2013phylogenetic}, LexStat distance, mean word length, distance between the
languages as features on cognate and non-cognate pairs extracted from word lists from
\citet{wichmann2013languages} and \citet{List2014d}. The details of the training dataset are given in table 1 in \citet{Jaeger2017}. We used the same training settings as reported in the paper to train our SVM model. The trained SVM model is then employed to compute the probability that a word pair is cognate or not. The word pair probability matrix is then given as input to InfoMap algorithm for inferring word clusters. The threshold for InfoMap algorithm is set to 0.57 after cross-validation experiments on the training data.

\begin{table*}[!ht]
\small \centering
\begin{tabular}{lccccc}
\toprule
Method & Austro-Asiatic & Austronesian & Indo-European & Pama-Nyungan & Sino-Tibetan\\
\midrule
CCM     &  	0.71 & 0.7 & 0.75 & 0.74 & 0.48 \\ 
NED      & 	0.73 & 0.77	& 0.69 & 0.53 & 0.49 \\
SCA       &	0.76 & 0.78 & 0.81 & 0.71 & 0.56 \\
LexStat   &	0.76 & \cellcolor{lightgray}0.84 & \cellcolor{lightgray}0.83 & 0.84 & \cellcolor{lightgray}0.6 \\
OnlinePMI &	0.76 & 0.81 & 0.82 & 0.72 & 0.56 \\
SVM  &     	\cellcolor{lightgray}0.82 & 0.81	& 0.79 & \cellcolor{lightgray}0.86 & 0.5 \\
\bottomrule
\end{tabular}
\caption{B-cubed F-scores for different cognate detection methods across the language families.}
\label{tab:bcubedfscores}
\end{table*}

We evaluate the quality of the inferred cognate sets using the above described methods using B-cubed F-score \citep{amigo2009comparison} which is widely used in evaluating the quality of automatically inferred cognate clusters \citep{hauer-kondrak:2011:IJCNLP-2011}. We present the cognate evaluation results in table \ref{tab:bcubedfscores}. The SVM system is the best in the case of Austro-Asiatic and Pama-Nyungan whereas LexStat algorithm performs the best in the case of rest of the datasets. This is surprising since LexStat scores are used as features for SVM and we expect the SVM system to perform better than LexStat in all the language families. On the other hand, both OnlinePMI and SCA systems perform better than the algorithmically simpler systems such as CCM and NED. Given these F-scores, we hypothesize that the cognate sets output from the best cognate identification systems would also yield the high quality phylogenetic trees. However, we find the opposite in our phylogenetic experiments.

\section{Bayesian Phylogenetic Inference}\label{sec:bayinf}
The objective of Bayesian phylogenetic inference is based on the Bayes rule in \ref{eq:bphy}.
\begin{equation}\label{eq:bphy}
f(\tau, v,\theta|X) = \frac{f(X|\tau, v,\theta)f(\tau, v,\theta)}{f(X)}
%{\sum_\tau \int_v \int_\theta f(X|\tau, v,\theta)f(\tau, v,\theta) dv d\theta }
\end{equation}
where $X$ is the data matrix, $\tau$ is the topology of the tree, $v$ is the vector of branch lengths, and $\theta$ is the substitution model parameters. The data matrix $X$ is a binary matrix of dimensions $N \times C$ where $N$ is the number of languages and $C$  is the number of cognate clusters in a language family. The posterior distribution $f(\tau, v, \theta|X)$ is difficult to calculate analytically since one has to sum over all the possible topologies ($\frac{(2N-3)!}{2^{N-2}(N-2)!}$) to compute the marginal in the denominator. However, posterior probability of all the parameters of interest (here, $ \Psi = \{\tau, v, \theta\}$) can be computed from samples drawn using a Markov chain Monte Carlo (MCMC) method. Typically, Metropolis-Hastings (MH) algorithm is the MCMC algorithm used to sample phylogenies from the  posterior distribution \citep{huelsenbeck2001bayesian}.

The MH algorithm constructs a Markov chain of the parameters' states by proposing change to a single parameter or a block of parameters in $\Psi$. The current state $\Psi$ in the Markov chain has a parameter $\theta$ and a new value $\theta^*$ is proposed from a distribution $q(\theta^*|\theta)$, then $\theta^*$ is accepted with a probability

\begin{equation}\label{eq:mhr}
r = \frac{f(X|\tau, v,\theta^*)}{f(X|\tau, v,\theta)}\frac{f(\theta^*)}{f(\theta)} \frac{q(\theta|\theta^*)}{q(\theta^*|\theta)}
\end{equation}
The likelihood of the data $f(X|\Psi)$ is computed using the Felsenstein's pruning algorithm \citep{felsenstein1981evolutionary} also known as sum-product algorithm \citep{jordan2004graphical}. We assume that $\tau, \theta, v$ are independent of each other.
%\footnote{A textbook length treatment of statistical phylogeny can be found in \citet{yang2014molecular}}

\section{Experiments}\label{sec:results}
In this section, we report the experimental settings, the evaluation measure, and the results of our experiments.
\begin{table*}[t]
\small \centering
\begin{tabular}{llllll}
\toprule
Method & Austro-Asiatic & Austronesian & Indo-European & Pama-Nyungan & Sino-Tibetan\\
\midrule
Expert cognate sets      & \bfseries 0.0081 $\pm$ 0.001              &0.1056 $\pm$ 0.0118                       &\bfseries 0.0249 $\pm$ 0.0079            &\bfseries 0.1384 $\pm$ 0.0225             &\bfseries 0.0561 $\pm$ 0.0123 \\\midrule
CCM       & 0.0243 $\pm$ 0.018                        &0.0854 $\pm$ 0.0176                       &0.0369 $\pm$ 0.0148                      &0.1617 $\pm$ 0.0162                       &0.1424 $\pm$ 0.027 \\
NED       & 0.0265 $\pm$ 0.007                        &\cellcolor{lightgray}0.0458 $\pm$ 0.0152 &0.046 $\pm$ 0.0132                       &0.196 $\pm$ 0.0166                        &0.1614 $\pm$ 0.0282 \\
SCA       &0.0152 $\pm$ 0.0035 &0.0514 $\pm$ 0.013                        &\cellcolor{lightgray}0.0256 $\pm$ 0.009 &0.166 $\pm$ 0.0153                        &\cellcolor{lightgray}0.0704 $\pm$ 0.0206 \\
LexStat   & 0.0267 $\pm$ 0.0085                       &0.0848 $\pm$ 0.0226                       & 0.0314 $\pm$ 0.0091                      &\cellcolor{lightgray}0.1507 $\pm$ 0.0143 &0.0786 $\pm$ 0.0209 \\
OnlinePMI & 0.0158 $\pm$ 0.0048                       &0.1056 $\pm$ 0.0198                       & 0.0457 $\pm$ 0.0135                      & 0.1717 $\pm$ 0.0185                       &0.1184 $\pm$ 0.031 \\
SVM       & \cellcolor{lightgray}0.0146 $\pm$ 0.0039                       &0.0989 $\pm$ 0.0224                       & 0.0452 $\pm$ 0.011                       & 0.1827 $\pm$ 0.0237                       &0.1199 $\pm$ 0.0269 \\
\bottomrule
\end{tabular}
\caption{The mean and standard deviation for each method and family is computed from 7500 posterior trees. The automatic methods which comes closest to the gold standard phylogeny is shaded in gray, and where the expert cognate sets perform best, this is indicated with a \textbf{bold} font.}
\label{tab:gqd}
\end{table*}

All our Bayesian analyses use binary datasets with states $0$ and $1$. We employ the Generalized Time Reversible Model \citep[chapter 1]{yang2014molecular} for computing the transition probabilities between individual states. The rate variation across sites is modeled using a four category discrete $\Gamma$ distribution \citep{yang1994maximum}. We follow \citet{lewis2001likelihood} and \citet{felsenstein1992phylogenies} in correcting the likelihood calculation for ascertainment bias resulting from unobserved \texttt{0} patterns. We used a uniform tree prior \citep{ronquist2012total} in all our analyses which constructs a rooted tree and draws internal node heights from uniform distribution. In our analysis, we assumes a Independent Gamma Rates relaxed clock model \citep{lepage2007general} where the rate for a branch $j$ of length $b_j$ in the tree is drawn from a Gamma distribution with mean 1 and variance $\sigma^2_{IG}/b_j$ where $\sigma^2_{IG}$ is a parameter sampled in the MCMC analysis.

We infer $\tau, v, \theta$ from two independent random starting points and sample every 1000th state in the chain until the phylogenies from the two independent runs do not differ beyond $0.01$. For each dataset, we ran the chains for 15 million generations and threw away the initial $50\%$ of the chain's states as part of burnin. After that we computed the generalized quartet distance from each of the posterior trees to the gold standard tree described in subsection \ref{subsec:gqd}. All our experiments are performed using MrBayes 3.2.6 \cite{zhang2015total}.

%We used MrBayes software \citep{ronquist2012mrbayes} in our experiments.

\subsection{GQD}\label{subsec:gqd}
%\textcolor{red}{Taraka can write this part on GQD}
%We compare the posterior distribution of trees to the Glottolog tree using Generalized Quartet Distance. 
\citet{pompei2011accuracy} introduced Generalized Quartet Distance (GQD) as an extension to Quartet
Distance (QD) in order to compare binary trees with a polytomous tree, since gold standard trees can have non-binary internal nodes. It was widely used for comparing inferred language phylogenies with gold standard phylogenies \citep{greenhill2010accurate,Wichmann:2011:2210-5824:205,jager2013phylogenetic}.

QD measures the distance between two trees in terms of the number of different quartets
\citep{estabrook1985comparison}. A quartet is defined as a set of four leaves selected from a set of
leaves without replacement. A tree with $n$ leaves has ${n \choose 4}$ quartets in total. A quartet
defined on four leaves $a,b,c,d$ can have four different topologies: $ab|cd$, $ac|bd$, $ad|bc$, and
$ab\times cd$. The first three topologies have an internal edge separating two pairs of leaves. Such
quartets are called as \emph{butterflies}. The fourth quartet has no internal edge and as such is
known as star quartet. Given a tree $\tau$ with $n$ leaves, the quartets can be partitioned into
sets of butterflies, $B(\tau)$, and sets of stars, $S(\tau)$. Then, the QD between $\tau$ and $\tau_g$ is defined as $1-\frac{|S(\tau)\cap S(\tau_g)|+|B(\tau)\cap B(\tau_g)|}{{n \choose 4}}$. The QD formulation counts the butterflies in an inferred tree $\tau$ as errors. The tree $\tau$ should not be penalized if an internal node in the gold standard tree $\tau_g$ is $m$-ary. To this end, \citet{pompei2011accuracy} defined a new measure known as GQD to discount the presence of star quartets in $\tau_g$. GQD is defined as $DB(\tau, \tau_g)/B(\tau_g)$ where $DB(.)$ is the number of butterflies between $\tau,  \tau_g$.

We extracted gold standard trees from Glottolog \citep{Hammarstroem2017} for the purpose of evaluating the inferred posterior trees from each automated cognate identification system. We note that the Bayesian inference procedure produces rooted trees with branch lengths whereas the gold standard trees do not have any branch lengths. Although there are other linguistic phylogenetic inference algorithms such as those of \citet{ringe2002indo} we do not test the algorithms due to the non-availability and scalability of the software to datasets with more than twenty languages.
%\subsection{Tests}
%\textcolor{red}{write here what tests we carried out and that the source code with longer descriptions is available at OSF}

\subsection{Results}
The results of our experiments are given in table \ref{tab:gqd}. A average lower GQD score implies that the inferred trees are closer to the gold standard phylogeny than a higher average GQD score. Except for Austronesian, Bayesian
inference based on expert cognate sets yields trees that are very close to the gold standard tree.
Surprisingly, algorithmically simple systems such as NED and CCM show better performance than the
machine-learned SVM model except from Sino-Tibetan. 
SCA is a subsystem of LexStat but emerges as the winner in two language families (Indo-European and Sino-Tibetan). 
Given that SCA is outperformed by SVM and LexStat in automatic cognate detection, this is very
surprising, and further research is needed to find out, why the simpler models perform well on
phylogenetic reconstruction.
Although our results indicate that expert-coded cognate sets are generally more suitable for
phylogenetic reconstruction, we can also see that the difference to trees inferred from
automated cognate sets is not very large.  

\section{Conclusion}\label{sec:concl}

In this paper, we carried out a preliminary evaluation of the usefulness of automated cognate
detection methods for phylogenetic inference. Although the cognate sets predicted by automated
cognate detection methods yield phylogenetic trees that come close to expert trees, there is still
room for improvement, and future research is needed to further enhance automatic cognate detection
methods. However, as our experiments show, expert-annotated cognate sets are also not free from
errors, and it seems likewise useful to investigate, how the consistency of cognate coding by
experts could be further improved.

As future work, we intend to create a cognate identification system that combines the output of different algorithms in a more systematic way. We intend to infer cognate sets from the combined system and use them to infer phylogenies and evaluate the inferred phylogenies against the gold standard trees.

{\begin{spacing}{1.2} 
\section*{Acknowledgments}
\footnotesize This research was supported by the ERC Advanced Grant 324246 EVOLAEMP (GJ, JW), the DFG-KFG 2237 Words,
Bones, Genes, Tools (GJ), the ERC Starting Grant 715618 CALC (JML), and the BIGMED project (TR).
We thank our anonymous reviewers for helpful comments,
Mei-Shin Wu for helping with Sino-Tibetan data, Claire Bowern and Tiago Tresoldi for helping with
Pama-Nyungan data, Paul Sidwell for helping with Austro-Asiatic data, as well as the audience at the
CESC 2017 conference (MPI-SHH, Jena) for their
helpful comments on an earlier version of the paper.\end{spacing}}

\bibliography{naaclhlt2018}
\bibliographystyle{acl_natbib}

\appendix

\section{Supplemental Material}\label{sec:supplemental}
The code and data used in this paper are uploaded as a zip file. In addition, they are available for
download via Zenodo at \url{https://doi.org/10.5281/zenodo.1218060}.

\end{document}